\documentclass[11pt]{article}

\usepackage[margin=1in]{geometry}
\usepackage{setspace}
\usepackage{amsmath,amssymb,amsfonts}
\usepackage{textcomp}

\usepackage{graphicx}
\usepackage{subcaption}
\usepackage{booktabs}
\usepackage[table]{xcolor}
\usepackage{multirow}
\usepackage{wrapfig}

\usepackage[numbers,sort&compress]{natbib}
\usepackage{hyperref}
\hypersetup{
    colorlinks=true,
    linkcolor=blue,
    citecolor=blue,
    urlcolor=blue
}

\graphicspath{{plots/}{plots/new_plots/}}

\title{Leveraging Transfer Learning with Class-Specific Decoders\\ for Laparoscopic Segmentation}

\author{
    Priya Tomar$^{1,2,*}$,~
    Aditya Parikh$^{1,*}$,~
    Christian Bauckhage$^{1,2}$,~
    Rafet Sifa$^{1,2}$\\[6pt]
    $^{1}$Fraunhofer IAIS \quad
    $^{2}$University of Bonn, Germany\\[4pt]
    \texttt{priya.priya@iais.fraunhofer.de, ppriya@uni-bonn.de,}\\
    \texttt{\{firstname.lastname\}@iais.fraunhofer.de}\\[4pt]
    {\small $^{*}$Contributed equally}
}
\date{}

\begin{document}
\maketitle
\vspace{-0.5cm}
\begin{center}
\small
\textit{Published in: 2025 IEEE International Conference on Big Data (BigData)}\\
\textit{DOI:} \href{https://ieeexplore.ieee.org/document/11401034}{10.1109/BigData62323.2025.11401034}\\[2pt]
\textcopyright~2025 IEEE. Personal use of this material is permitted. Permission from IEEE must be obtained for all other uses.
\end{center}
\vspace{0.3cm}

\begin{abstract}
Effective multi-organ segmentation in surgical data requires learning the intricate anatomical features and alleviating the challenge of class imbalance, which results from relatively lower proportions of small and limitedly exposed structures. Recent works on laparoscopic multi-organ segmentation focus on learning structure-specific features through class-specific decoder architectures and report favorable results. This work extends the decoder-focused architectures to investigate knowledge sharing in the cross-surgical domain. We utilize two datasets representing different surgical domains, rectal and cholecystectomy surgeries, to explore how surgical conceptual knowledge transfers under partially common anatomical representations. Additionally, we compare the feature adaptation for the encoder and decoder at different training stages to analyse the knowledge adaptation and retention in the network. Our results corroborate previous findings on decoder-specific architectures and demonstrate that the organ-specific decoder model (CEMD), fully fine-tuned after cross-domain pre-training, achieves the highest segmentation performance (62.4\% dice) while converging substantially faster than training from scratch. However, we also find that class imbalance in surgical data remains a persistent challenge that transfer learning does not fully resolve for underrepresented anatomical structures.
\end{abstract}

\noindent\textbf{Keywords:} Surgical Segmentation, Multi-Organ Segmentation, Transfer Learning, Laparoscopic surgery, Dresden Surgical Anatomy dataset, CholecSeg8K

\section{Introduction}
The application of deep learning in the surgical domain has resulted in various developments concerning surgical phase detection, action recognition, lesion detection, detection of anatomical structures, anomaly classification, and semantic segmentation, etc.\ \cite{zhou2025vision}. In surgical segmentation tasks, extensive focus has been given to instrument segmentation and tracking \cite{rueckert2024methods,ahmed2024deep,wu2025augmenting,wei2025segmatch}. Existing research on surgical organ segmentation primarily involves organ-specific segmentation due to the limited availability of pertinent multi-organ datasets \cite{para11,para14}. Beyond challenges in data processing and ethical considerations, it requires domain expertise to annotate anatomical boundaries accurately under complex surgical conditions, including variable lighting, inconsistent organ appearance, instrument occlusions, and the presence of bodily fluids \cite{para13, ch8k_split, para22}.

Recently developed multi-organ surgical segmentation datasets have facilitated the investigation of deep learning approaches for surgical scene segmentation \cite{intro2,hong2020cholecseg8k,article_dsad}. The Dresden Surgical Anatomy (DSA) dataset \cite{article_dsad} is the biggest laparoscopic segmentation dataset, subsuming 13195 weakly-annotated images, and CholecSeg8K \cite{hong2020cholecseg8k}, containing 8080 images, has the highest number of densely-annotated classes. Various factors, including spatial and visual attributes of classes, influence the segmentation results. The correlation of organ representation in terms of pixel area with segmentation performance is highly evident and reflected by the best performance on organs with relatively high visual proportions like the Abdominal Wall and Liver \cite{kolbinger2024strategies, intro1, dsad1, tomar2025effective}. Smaller organs like the Spleen have better results due to the distinctive visual features than adjacent classes. This class imbalance results in poor learning of underrepresented classes, hence limiting the clinical applicability.

Recent works have investigated the efficacy of class-specific modules to learn anatomy-specific features for segmentation on laparoscopic images \cite{intro1, dsad1, dsad2, tomar2025effective}. \cite{tomar2025effective} has demonstrated the effectiveness of organ-specific decoder training in comparison to the single decoder network of comparable size, for five different architectures. Furthermore, they explored disjoint learning strategies where gradient updates are selectively applied to decoders corresponding to anatomical classes present in training samples, demonstrating computational efficiency while maintaining segmentation performance.

Transfer learning methodologies have gained considerable traction in medical image segmentation, particularly in scenarios characterized by limited annotated data availability, which is also a persistent challenge in surgical imaging domains \cite{liu2024towards,chen2021transunet}. In this work, we further leverage the CholecSeg8K dataset for learning surgical anatomical context in a supervised learning approach. The intersection of these datasets presents a unique opportunity for knowledge transfer, leveraging shared anatomical representations while accommodating procedural variations between the involved surgical categories. This work extends \cite{tomar2025effective}, which studied organ-specific decoder architectures trained from scratch on the DSA dataset only. In contrast, we introduce three novel dimensions: (i)~evaluation of class-specific decoders on a second surgical dataset (CholecSeg8K), (ii)~cross-surgical-domain transfer learning between cholecystectomy and rectal surgery data, and (iii)~a systematic comparison of decoder-only versus full-network fine-tuning to disentangle encoder and decoder contributions during domain adaptation. Our specific contributions are:
\begin{itemize}
    \item To the best of our knowledge, we present the first evaluation of organ-specific decoder learning (CEMD) on CholecSeg8K, showing that CEMD yields an overall dice gain of approximately 4\% over the shared-decoder baseline (CECD), with the benefit concentrated on the underrepresented Gallbladder class ($+$27\% dice), reinforcing the value of class-specific decoders for imbalanced surgical data.
    \item We investigate cross-surgical domain transfer from cholecystectomy (CholecSeg8K) to rectal surgery (DSA) data under partially shared anatomical classes. Without any fine-tuning, the transferred CEMD model achieves an overall dice score of approximately 60\% on DSA, comparable to training from scratch, demonstrating effective knowledge transfer across surgical procedures.
    \item We compare decoder-only fine-tuning (DFT) and full-network fine-tuning (FFT) to isolate encoder versus decoder adaptation. DFT consistently degrades performance, while FFT improves it, with CEMD\textsubscript{FFT} achieving the highest overall dice (62.4\%). Notably, CEMD is more resilient to DFT degradation than CECD ($-$4.9\% vs.\ $-$5.6\% overall dice), indicating that organ-specific decoders retain transferred knowledge better under domain shift.
\end{itemize}

\section{Related Work}

\subsection{Multi-Organ Segmentation in Surgical Imaging}
Multi-organ segmentation in laparoscopic surgery poses unique challenges compared to radiology-based tasks, including variable lighting, deformable anatomy, and instrument occlusions~\cite{zhou2025vision, qi2025advancements}. The DSA dataset~\cite{article_dsad} and CholecSeg8K~\cite{hong2020cholecseg8k} have enabled systematic benchmarking, with several works proposing architecture-level solutions. Kolbinger~et~al.~\cite{kolbinger2024strategies} investigated strategies for improving real-world applicability, while Jenke~et~al.~\cite{dsad1} explored training a single model using complementary, partially labeled datasets. Maack~et~al.~\cite{dsad2} proposed multi-teacher knowledge distillation, leveraging anatomy-specific teacher networks to improve a compact student model. Urrea~et~al.~\cite{urrea2024improving} addressed class imbalance through architectural attention mechanisms. Our work complements these efforts by investigating decoder-specific architectures across datasets and surgical domains rather than within a single dataset.

\subsection{Transfer Learning in Surgical Computer Vision}
Transfer learning from ImageNet pre-trained models is standard practice in medical imaging~\cite{tf1,tf2,tf3}, yet its utility in surgical vision has been questioned. Raghu~et~al.~\cite{tf2} showed that ImageNet initialization offers limited performance gains for medical tasks while primarily accelerating convergence. More recently, domain-specific pre-training has gained attention: Alapatt~et~al.~\cite{alapatt2024jumpstarting} demonstrated that the composition of surgical pre-training datasets critically influences downstream task performance through self-supervised learning. Jaspers~et~al.~\cite{jaspers2025surgenet} scaled this approach with SurgeNetXL, a foundation model pre-trained on over 4.7 million surgical video frames, achieving state-of-the-art results across segmentation, phase recognition, and safety classification tasks. Unlike these self-supervised approaches, our work investigates supervised cross-domain transfer between two labeled surgical datasets with partially overlapping anatomical classes, focusing on how encoder and decoder components differentially adapt under domain shift.

\subsection{Class-Specific and Multi-Decoder Architectures}
The idea of dedicating separate decoder pathways to individual classes or tasks has been explored across medical imaging. In the surgical domain, Tomar~et~al.~\cite{tomar2025effective} systematically compared organ-specific decoders against shared-decoder architectures across five backbones on the DSA dataset, demonstrating consistent improvements for class-specific decoders particularly on underrepresented organs. Wang~et~al.~\cite{wang2025efficient} proposed a generative-adversarial U-Net with specialized decoding branches for multi-organ segmentation. The class imbalance problem underlying decoder-specific designs has received broad attention: Xu~et~al.~\cite{xu2025balance} proposed a unified loss function addressing both inter-class and intra-class imbalance in medical segmentation. Our work extends the decoder-specific paradigm~\cite{tomar2025effective} in two directions, applying it to a second surgical dataset (CholecSeg8K) and investigating its interaction with cross-domain transfer learning.

\section{Methods \& Experiments}

\subsection{Data}
We consider two surgical multi-organ segmentation datasets in our work:
\begin{enumerate}
    \item \textit{Dresden Surgical Anatomy (DSA) Dataset}: It is a laparoscopic segmentation dataset, containing 13195 images and semantic segmentation masks from 32 rectal surgeries. It provides binary masks for eleven anatomical classes, including eight abdominal organs (Colon, Liver, Pancreas, Small Intestine, Spleen, Stomach, Ureter, and Vesicular Glands), the Abdominal Wall, and two vessel structures (Inferior Mesenteric Artery and Intestinal Veins). Each anatomical class has at least 1000 samples annotated by three surgeons. Besides, it includes a multi-class subset of 1430 images with dense annotations containing masks for all visible organs in the image. The DSA dataset is pre-processed for relevant frame selection and trimming before annotation by three surgeons. For detailed information concerning the dataset, refer to \cite{article_dsad}. We use the binary segmentation subset of DSA in our experiments.
    \item \textit{CholecSeg8K Dataset}: It subsumes 8080 images of laparoscopic cholecystectomy surgeries, extracted from the Cholec80 dataset \cite{twinanda2016endonet}. Apart from background, it contains multi-class masks for twelve classes, including abdominal wall, liver, gastrointestinal tract, fat, grasper, connective tissue, blood, cystic duct, L-hook electrocautery, gallbladder, hepatic vein, and liver ligament. In addition to anatomical classes, CholecSeg8K contains two instrument classes (grasper and L-hook electrocautery). A subset of relevant videos from the Cholec80 dataset is extracted at the original frame rate (25 frames per second) and densely annotated. Each patient video contains clips of 80 frames, making the frames in a clip visually similar to the adjacent frames. Refer to \cite{hong2020cholecseg8k} for additional details on this dataset.
\end{enumerate}

DSA and CholecSeg8K datasets share anatomical context. They have two common annotated classes: Liver and Abdominal Wall. Besides, the gastrointestinal class in CholecSeg8K combines stomach, small intestine, and nearby tissues, which represent distinctive classes in the DSA dataset.

\subsection{Training Framework}

\begin{figure}[htb]
    \centering
    \includegraphics[width=0.65\linewidth]{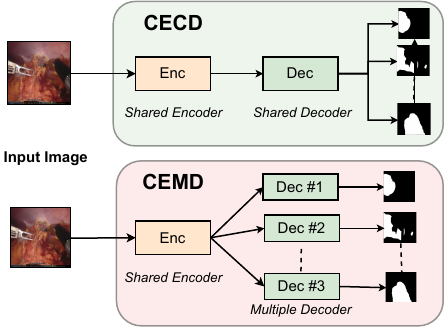}
    \caption{Comparison of (top) Common Encoder-Common Decoder (CECD): A single pipeline processes input images, producing multi-channel output corresponding to organ segmentation masks. This approach shares features across organs while maintaining efficiency. (bottom) Common Encoder-Multiple Decoder (CEMD): This architecture uses a shared encoder with dedicated decoders for each of the eleven target organs. The common encoder captures general abdominal anatomy, while specialized decoders focus on organ-specific features.}
    \label{fig:cecd_cemd}
\end{figure}

Considering the benefits of organ-specific decoders, we investigate two training frameworks in our investigation, namely Common Encoder Common Decoder (CECD) and Common Encoder Multiple Decoder (CEMD). We present the illustration of CECD and CEMD in Figure~\ref{fig:cecd_cemd}.
\begin{itemize}
    \item \textit{Common Encoder Common Decoder}: As the name suggests, the CECD approach processes the encoder representation through a single decoder to produce multi-channel masks respective to each organ, enabling the decoder to learn the cumulative representation of input classes and the surgical setting.
    \item \textit{Common Encoder Multiple Decoder}: CEMD focuses on class-oriented learning by using a common encoder but a separate decoder for each class. It promotes class-specific feature learning.
\end{itemize}

We use the Attention U-Net \cite{aunet} model for evaluating CECD and CEMD frameworks, considering the computational requirements and results reported by \cite{tomar2025effective} after investigating different training architectures.

In aggregate, we evaluate three different training strategies in our experiments, one for training from scratch and two transfer learning approaches concerning the decoder adaptation:
\begin{enumerate}
    \item \textit{No Fine-Tuning (CECD\textsubscript{0FT}, CEMD\textsubscript{0FT})}: For in-dataset evaluation, the model is trained from scratch with random weight initializations and evaluated on the same dataset. For cross-dataset evaluation, the model trained on CholecSeg8K is directly evaluated on the DSA dataset without any fine-tuning.
    \item \textit{Decoder Fine-Tuning (CECD\textsubscript{DFT}, CEMD\textsubscript{DFT})}: We pre-train the model on CholecSeg8K and fine-tune on the DSA dataset by freezing the parameters of the encoder and only updating the decoder during fine-tuning, to isolate and assess the architectural dependency on encoder and decoder features.
    \item \textit{Full Fine-Tuning (CECD\textsubscript{FFT}, CEMD\textsubscript{FFT})}: After pre-training on CholecSeg8K, the entire network is fine-tuned on the DSA dataset.
\end{enumerate}

\subsection{Experiment Setup}
\label{appendix:setup}
We initialize all models with a fixed seed and identical configuration parameters for reproducibility and fair comparison. Models are trained for a maximum of 150 epochs using dice loss and the Adam optimizer with an initial learning rate of $1 \times 10^{-4}$, which is decreased by a factor of 0.5 every 10 epochs, and early stopping with a step size of 10 epochs.

We select the best model checkpoint for each architectural framework based on the lowest validation loss and report the results on the held-out test set. Besides, we save model checkpoints at 50, 100, and 150 epochs to evaluate the efficiency and convergence of our transfer learning approach compared to training from scratch. All experiments are conducted on NVIDIA A100 GPUs with the PyTorch framework.

\begin{figure}[!ht]
    \centering
    \includegraphics[width=0.65\linewidth]{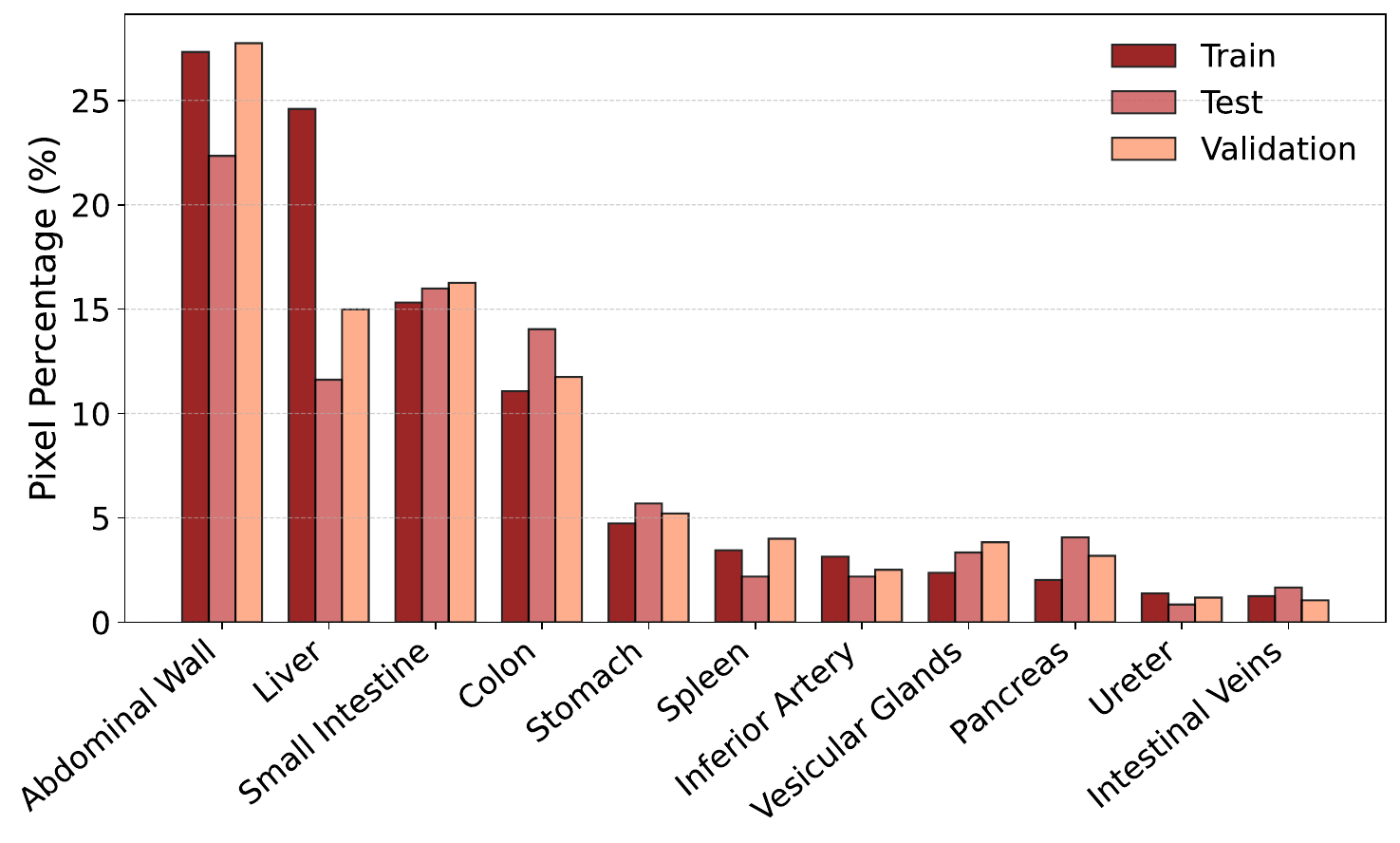}
    \caption{Pixel distribution of Dresden Surgical Anatomy (DSA) dataset classes in training and evaluation splits for the binary segmentation set. The classes are sorted according to their pixel percentage in the train set. The test split proportions are kept high and close to the train set for a reliable evaluation under class imbalance.}
    \label{fig:dsad_split_distribution}
\end{figure}

\begin{figure}[!ht]
    \centering
    \includegraphics[width=0.65\linewidth]{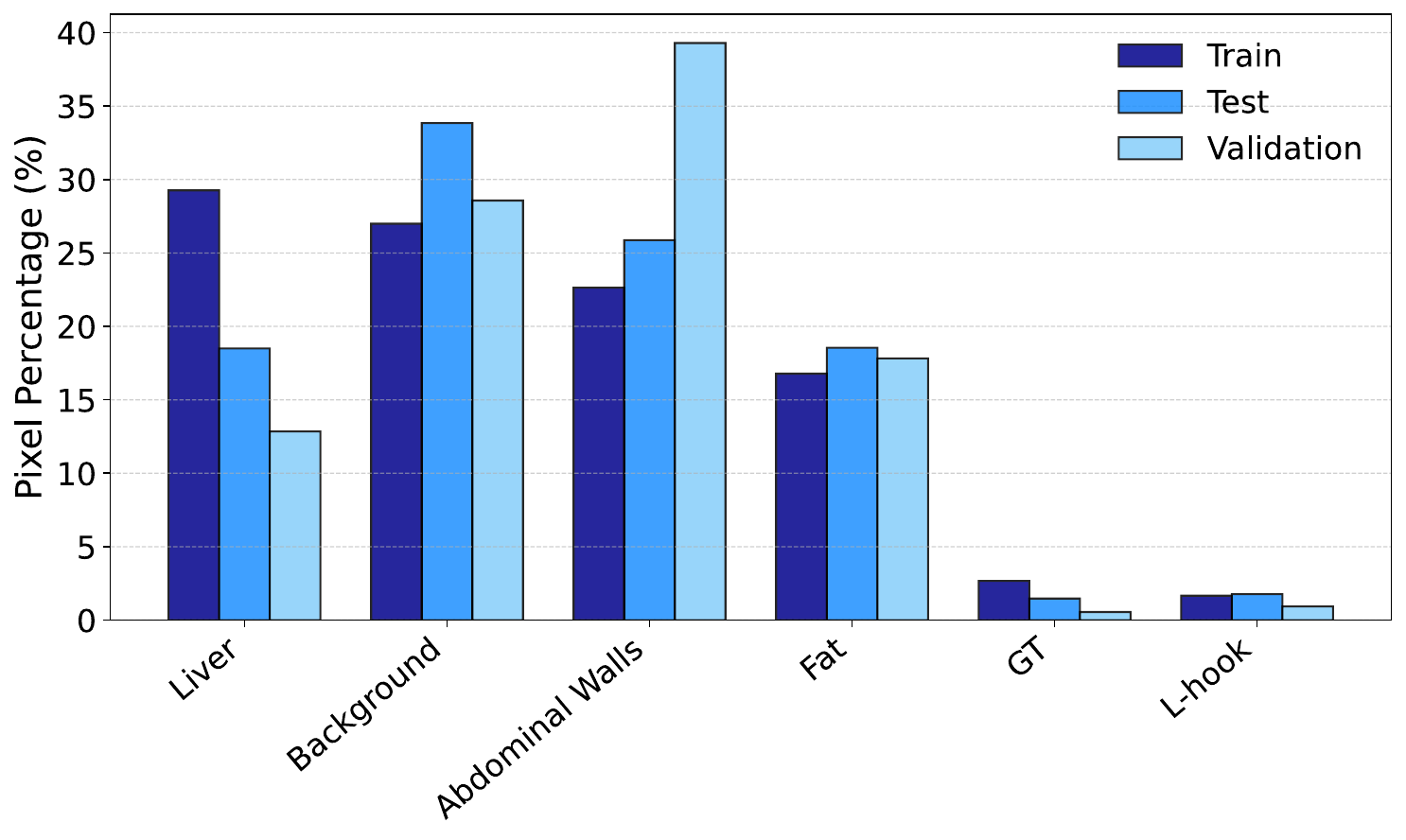}
    \caption{Pixel distribution of CholecSeg8K dataset classes in training and evaluation splits. The classes are sorted according to their pixel percentage in the train set. GT represents the Gallbladder class. The test split proportions are kept high and close to the train set for a reliable evaluation under class imbalance.}
    \label{fig:cholec_split_distribution}
\end{figure}

We maintain dataset splits consistent with prior works \cite{tomar2025effective, intro1, ch8k_split}. Figure~\ref{fig:dsad_split_distribution} and Figure~\ref{fig:cholec_split_distribution} present the pixel-percentages of classes in training and evaluation splits for the DSA dataset and CholecSeg8K, respectively. For both datasets, we tried to keep the proportion of the test data high and within the range of the training data. This helps us obtain reliable results and account for the class imbalance during the analysis. Considering annotated classes, the Abdominal Wall and Liver are common classes in both datasets, with the highest and second-highest pixel area in the training sets. We evaluate segmentation performance using Dice Score (DS) and Intersection over Union (IoU).

\section{Results}
The following sections present the Dice score and IoU metrics obtained in the case of training from scratch (0FT) and two transfer learning, namely decoder fine-tuning (DFT) and full fine-tuning (FFT) approaches. We use the Dice score for the primary discussion on results.

\subsection{Training from scratch}

\subsubsection{DSA Dataset}

\begin{figure}[!hbt]
    \centering
    \includegraphics[width=0.65\linewidth]{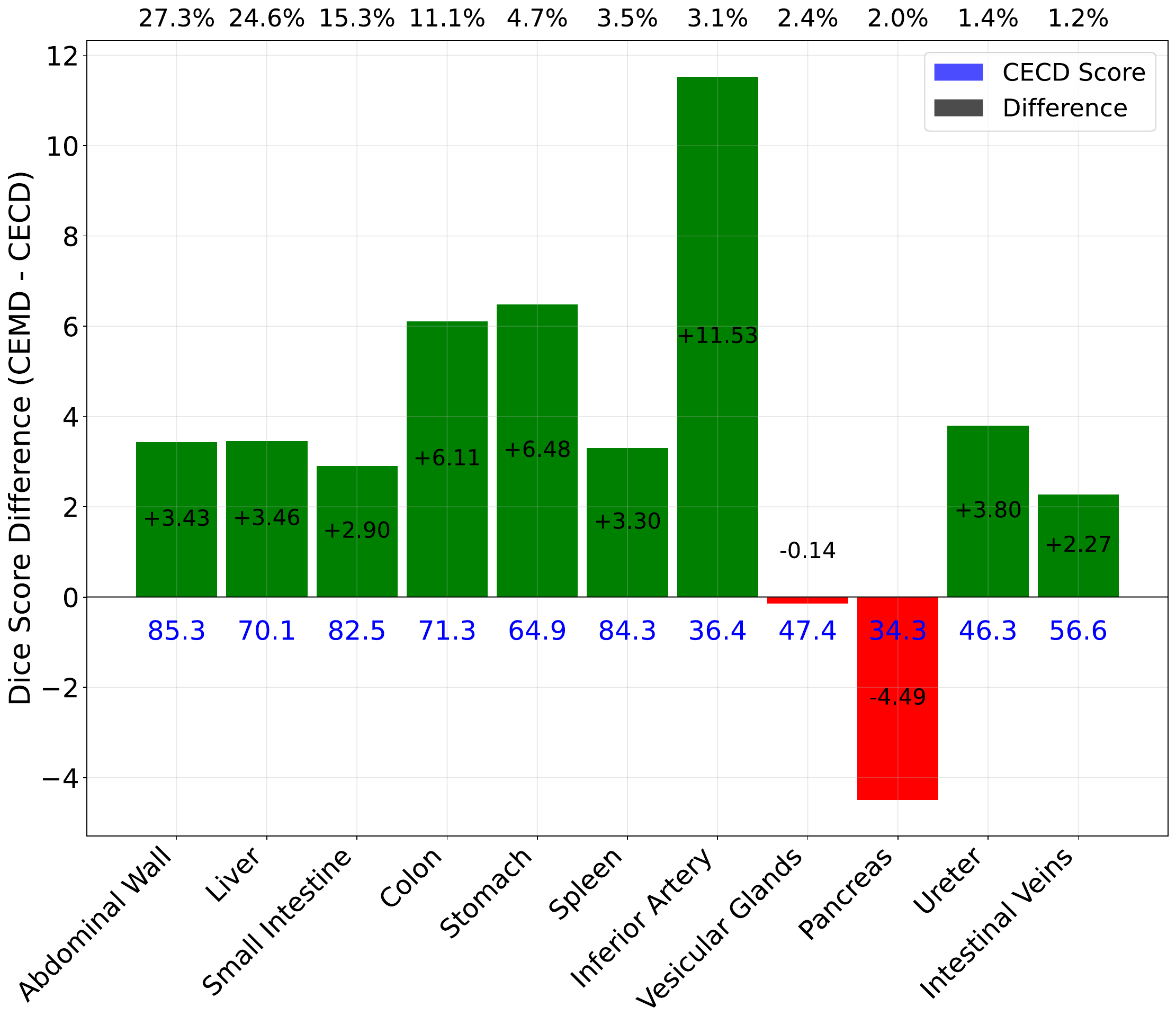}
    \caption{The difference in dice scores between the Common Encoder Multiple Decoder (CEMD) and Common Encoder Common Decoder (CECD) frameworks on the DSA dataset. The baseline indicates the CECD dice scores, and the pixel areas in the train set are shown on the top, sorted from the highest to the lowest pixel proportion in the training data.}
    \label{fig:cecd_cemd_dsa}
\end{figure}

Figure~\ref{fig:cecd_cemd_dsa} presents the performance comparison between the Common Encoder Common Decoder (CECD) and Common Encoder Multiple Decoder (CEMD) frameworks on the DSA dataset for each organ when models are trained from scratch. We mainly intend to deliberate over the effectiveness of CEMD on the DSA dataset by delineating the differences in dice scores, as a detailed analysis is presented in \cite{tomar2025effective}. In aggregate, CEMD improved scores for 9 out of 11 classes, with an average gain of about 3.5\% in the overall dice score and the highest increase of more than 10\% in the Inferior Mesenteric Artery class, indicating the advantage of organ-specific decoder learning. In addition to the visual characteristics of the anatomical classes, their pixel proportions influence the respective performances. That is, organs with higher pixel proportions (for instance, Abdominal Wall, Liver, and Small Intestine) have higher scores in general, whereas lower representation (for example, Pancreas, Inferior Mesenteric Artery, Ureter) translates to less than random predictions. Spleen and Intestinal Veins reflect comparatively improved scores despite their smaller size due to their conspicuously distinctive boundaries compared to adjacent anatomical areas.

\subsubsection{CholecSeg8K Dataset}
Out of all thirteen classes in CholecSeg8K, we have selected six classes, including Background, Abdominal Wall, Liver, Fat, Gallbladder, and L-hook electrocautery classes. The selected classes contain four anatomical and one instrument class. It is to provide preliminary and reliable insights into the CECD and CEMD paradigms, as the remaining classes have sparse representations. We present the scores on the comparison of common decoder (CECD) and organ-specific decoder (CEMD) training in Table~\ref{tab:cholecseg}.

\begin{table}[!htbp]
\centering
\caption{Dice score (DS) and IoU comparison between CECD and CEMD on CholecSeg8K (in \%). Organs are sorted according to the pixel \% in the train set. The pixel ratio of Background, Abdominal Wall, Liver, and Fat classes is more than 20\%, and the scores indicate comparatively better performance than for the underperforming classes Gallbladder (GT) and L-hook. The symbol * indicates a gain of more than 5\% for an organ in CEMD in comparison to CECD training, which is reflected in the Gallbladder class.}
\label{tab:cholecseg}
\small
\begin{tabular}{llcccc}
\toprule
\textbf{Organ} & \textbf{Pixel \%} & \multicolumn{2}{c}{\textbf{CECD}} & \multicolumn{2}{c}{\textbf{CEMD}} \\
\cmidrule(lr){3-4} \cmidrule(lr){5-6}
 & & DS & IoU & DS & IoU \\
\midrule
Liver          & 29.31 & 81.24 & 69.19 & 82.15 & 70.71 \\
Background     & 26.95 & 96.99 & 94.37 & 95.34 & 91.30 \\
Abdominal Wall & 22.64 & 88.88 & 80.46 & 86.31 & 76.67 \\
Fat            & 16.77 & 92.13 & 85.53 & 92.22 & 85.70 \\
Gallbladder    & 2.67  & 0.00  & 0.00  & 27.39* & 18.84* \\
L-hook         & 1.66  & 39.10 & 32.65 & 38.90 & 32.24 \\
\midrule
\textbf{Overall} & & 66.39 & 60.37 & \textbf{70.38} & \textbf{62.58} \\
\bottomrule
\end{tabular}
\end{table}

Comparing the organ-wise performances, both frameworks CECD and CEMD reflect the same performance trend, and this trend broadly follows the pixel proportion of the classes in the training data. When the scores are sorted from the highest to the lowest (Background, Fat, Abdominal Wall, Liver, L-hook electrocautery, and Gallbladder), the classes with a larger share in the training data generally achieve higher scores. The overall performance depends on the class attributes and the respective pixel percentages in training data, reflecting two distinguished categories: L-hook electrocautery and Gallbladder, with less than 3\% proportion, are the least performing in comparison to the remaining four classes. If the organ representation is sufficient for learning its features, then the visual characteristics are dominant in determining segmentation outcomes. For instance, notwithstanding the highest pixel proportion of the Liver class in the train set ($\sim$29\%), its scores are lower than the Fat class ($\sim$16\%).

With respect to the two training frameworks, CEMD in aggregate offers an overall increase in dice scores of $\sim$4\%, which can be mainly attributed to the considerable gain of approximately 27\% dice score in the Gallbladder class. For the remaining classes, the scores do not align with the performance trend on the DSA dataset, where CEMD outperformed CECD in nine out of eleven anatomical classes. Except for the Gallbladder class, CEMD results in either a marginal decrease (Background, Abdominal Wall) or an increase (Liver, Fat, L-hook electrocautery) in the dice scores.

\begin{wrapfigure}[17]{r}{0.49\linewidth}
    \centering
    \includegraphics[width=\linewidth]{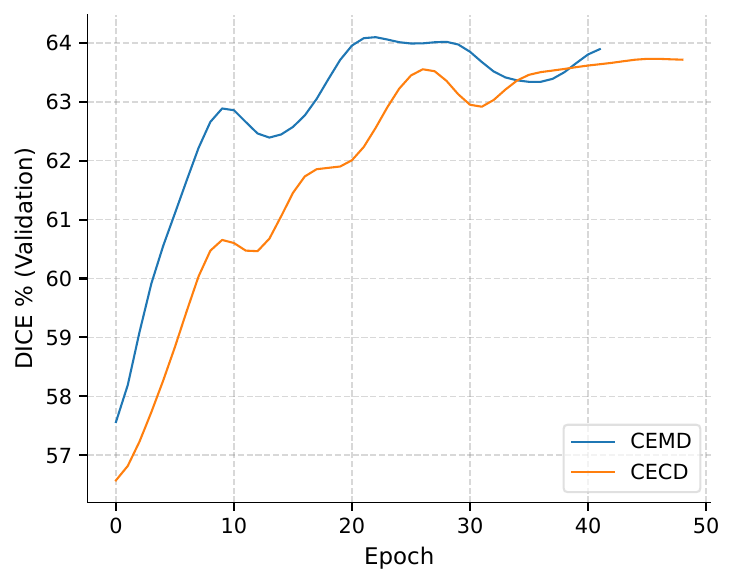}
    \caption{Comparison of DICE scores (\%) on the validation set across training epochs for the CholecSeg8K dataset using two architectural approaches, CECD and CEMD. Besides resulting in better performance, the class-specific decoder framework converges faster than the common decoder framework CECD.}
    \label{fig:cholecseg}
\end{wrapfigure}

We compare the learning curve of the two frameworks, CECD and CEMD, on a validation set at different training stages and plot the dice scores in Figure~\ref{fig:cholecseg}. Besides reflecting better aggregated performance, the plots delineate the early convergence of CEMD in comparison to CECD. The dedicated decoders for each class enable learning the representative class features faster.

\subsection{Knowledge Transfer}

\subsubsection{Segmentation Performance}
\begin{wraptable}{r}{0.49\textwidth}
\centering
\caption{Test set performance comparison (Dice Score, DS) across architectures and training approaches. CECD\textsubscript{0FT}/CEMD\textsubscript{0FT} indicates training on the CholecSeg8K dataset and direct evaluation on the DSA dataset without any fine-tuning, and it serves as the baseline. CECD\textsubscript{FFT}/CEMD\textsubscript{FFT} indicates full fine-tuning of the entire network on the DSA dataset, and CECD\textsubscript{DFT}/CEMD\textsubscript{DFT} denotes fine-tuning only the decoder units without updating the encoder parameters.}
\label{tab:results}
\small
\begin{tabular}{lcccc}
\toprule
\multirow{2}{*}{Architecture} & \multicolumn{3}{c}{Training Epochs} & \multirow{2}{*}{Best (\%)} \\
\cmidrule(lr){2-4}
 & 50 & 100 & 150 & \\
\midrule
CECD\textsubscript{0FT} & 39.7 & 52.4 & 56.5 & 59.3 \\
CECD\textsubscript{DFT} & \textbf{45.8} & 51.0 & 50.7 & 53.7 \\
CECD\textsubscript{FFT} & 33.4 & \textbf{53.0} & \textbf{57.3} & \textbf{59.4} \\
\midrule
\rowcolor[gray]{0.93} CEMD\textsubscript{0FT} & 58.0 & 60.4 & 60.7 & 60.0 \\
\rowcolor[gray]{0.93} CEMD\textsubscript{DFT} & 54.4 & 54.4 & 54.7 & 55.1 \\
\rowcolor[gray]{0.93} CEMD\textsubscript{FFT} & \textbf{59.9} & \textbf{61.8} & \textbf{62.4} & \textbf{62.4} \\
\bottomrule
\end{tabular}
\end{wraptable}
Table~\ref{tab:results} reports the results of knowledge transfer in the shared-decoder approach (CECD) and organ-specific multiple-decoder architecture (CEMD) from the CholecSeg8K to the DSA dataset. We consider no fine-tuning (CECD\textsubscript{0FT} vs.\ CEMD\textsubscript{0FT}) scores as the baseline for discussion on the transfer learning outcomes.

Firstly, consistent with \cite{tomar2025effective}, the overall dice scores demonstrate superior performance of CEMD compared to CECD in all three training approaches: no fine-tuning (CECD\textsubscript{0FT} vs.\ CEMD\textsubscript{0FT}) and the two fine-tuning approaches (CECD\textsubscript{DFT} vs.\ CEMD\textsubscript{DFT} and CECD\textsubscript{FFT} vs.\ CEMD\textsubscript{FFT}).

\begin{figure}[!htbp]
    \centering
    \begin{subfigure}[b]{0.7\textwidth}
        \centering
        \includegraphics[width=\linewidth]{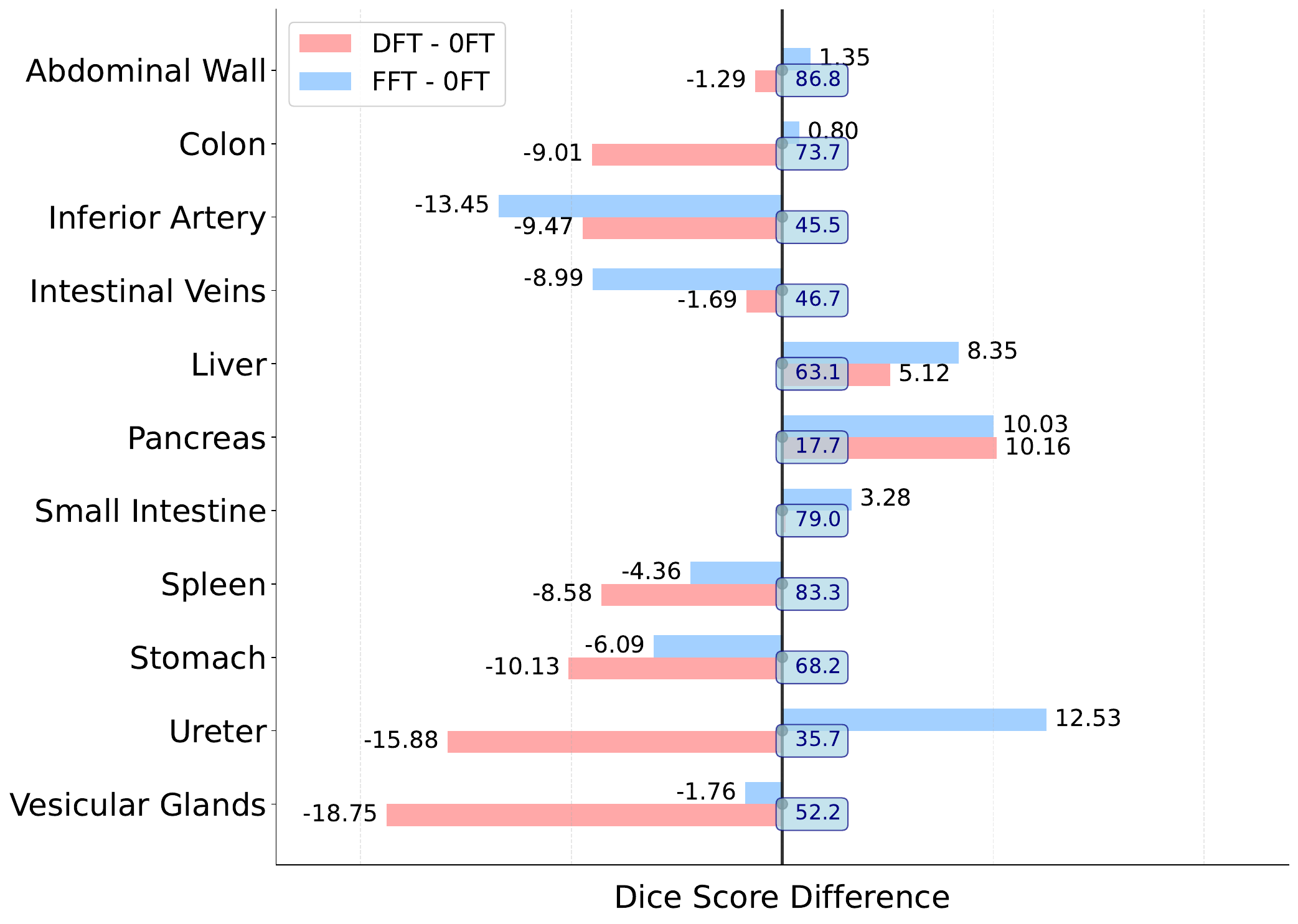}
        \caption{CECD}
        \label{fig:cecd_dsa_transfer_learning}
    \end{subfigure}

    \vspace{0.3cm}

    \begin{subfigure}[b]{0.7\textwidth}
        \centering
        \includegraphics[width=\linewidth]{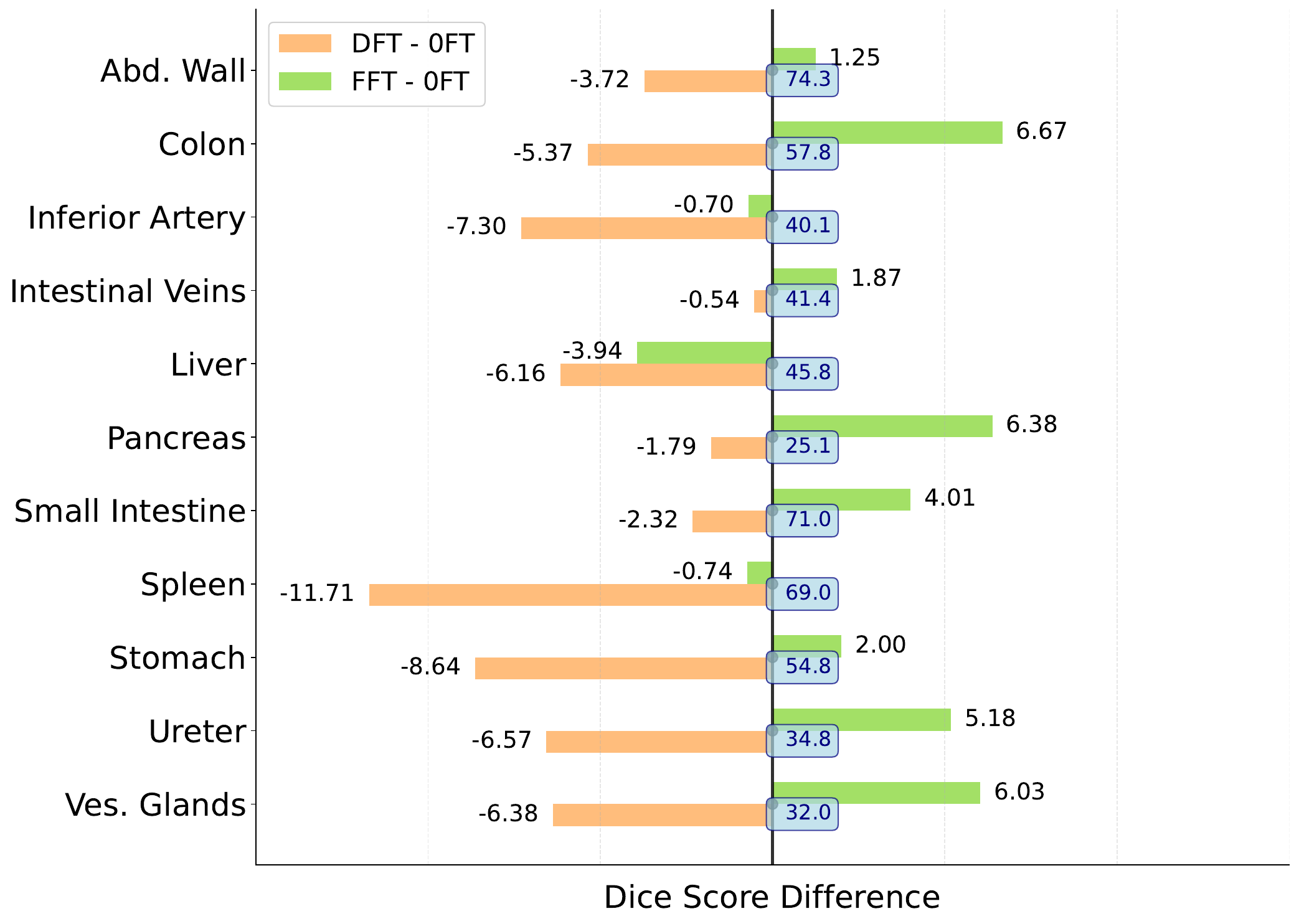}
        \caption{CEMD}
        \label{fig:cemd_dsa_transfer_learning}
    \end{subfigure}
    \caption{Difference in the dice scores for the Common Encoder Common Decoder (CECD) and Common Encoder Multiple Decoder (CEMD) training paradigms under the direct transfer or no fine-tuning approach (0FT), decoder fine-tuning (DFT), and full network fine-tuning (FFT) approaches, considering the 0FT dice score (in \%) as the baseline (middle) for each paradigm.}
    \label{fig:dice_transfer_learning}
\end{figure}

\begin{table*}[!htbp]
\centering
\caption{Performance comparison of transfer learning approaches for the Common-Encoder-Common-Decoder (CECD) architecture on the DSA dataset. $+$/$-$ denote an increase/decrease in the Dice (DS) or Intersection over Union (IoU) scores respectively in decoder fine-tuning (DFT) or full fine-tuning (FFT) in comparison to no fine-tuning (0FT) scores. 0FT indicates baseline scores for direct transfer learning from the CholecSeg8K dataset to the DSA dataset without any fine-tuning.}
\label{tab:cecd}
\small
\begin{tabular}{lcc|cc|cc}
\cmidrule(lr){2-7}
 & \multicolumn{2}{c}{No Fine-tuning (0FT)} & \multicolumn{2}{c}{Decoder Fine-tuning (DFT)} & \multicolumn{2}{c}{Full Fine-tuning (FFT)} \\
\cmidrule(lr){2-3} \cmidrule(lr){4-5} \cmidrule(lr){6-7}
Organ & Dice & IoU & Dice & IoU & Dice & IoU \\
\midrule
Abdominal Wall       & 86.81 & 77.17 & $-$85.52 & $-$75.13 & \textbf{$+$88.16} & \textbf{$+$79.12} \\
Colon                & 73.67 & 59.78 & $-$64.66 & $-$48.98 & \textbf{$+$74.47} & \textbf{$+$60.04} \\
Inferior Mesenteric  & \textbf{45.49} & \textbf{30.66} & $-$36.02 & $-$22.80 & $-$32.04 & $-$20.02 \\
Intestinal Veins     & \textbf{46.68} & \textbf{32.17} & $-$44.99 & $-$30.38 & $-$37.69 & $-$24.54 \\
Liver                & 63.12 & 49.36 & $+$68.24 & $+$54.36 & \textbf{$+$71.47} & \textbf{$+$58.67} \\
Pancreas             & 17.70 & 10.61 & \textbf{$+$27.86} & \textbf{$+$16.97} & $+$27.73 & $+$17.72 \\
Small Intestine      & 78.95 & 66.25 & $+$79.10 & $+$66.32 & \textbf{$+$82.23} & \textbf{$+$70.64} \\
Spleen               & \textbf{83.32} & \textbf{72.39} & $-$74.74 & $-$61.16 & $-$78.96 & $-$67.01 \\
Stomach              & \textbf{68.23} & \textbf{54.00} & $-$58.10 & $-$42.63 & $-$62.14 & $-$47.67 \\
Ureter               & 35.68 & 23.68 & $-$19.80 & $-$11.58 & \textbf{$+$48.21} & \textbf{$+$33.58} \\
Vesicular Glands     & \textbf{52.18} & \textbf{36.74} & $-$33.43 & $-$21.19 & $-$50.42 & $-$35.32 \\
\midrule
\textbf{Overall}     & 59.31 & 46.69 & $-$53.67 & $-$40.82 & \textbf{$+$59.44} & \textbf{$+$46.73} \\
\bottomrule
\end{tabular}
\end{table*}

\begin{table*}[!htbp]
\centering
\caption{Performance comparison of transfer learning approaches for the Common-Encoder-Multiple-Decoder (CEMD) architecture on the DSA dataset. $+$/$-$ denote an increase/decrease in the Dice (DS) or Intersection over Union (IoU) scores respectively in decoder fine-tuning (DFT) or full fine-tuning (FFT) in comparison to no fine-tuning (0FT) scores. 0FT indicates baseline scores for direct transfer learning from the CholecSeg8K dataset to the DSA dataset without any fine-tuning.}
\label{tab:cemd}
\small
\begin{tabular}{lcc|cc|cc}
\cmidrule(lr){2-7}
 & \multicolumn{2}{c}{No Fine-tuning (0FT)} & \multicolumn{2}{c}{Decoder Fine-tuning (DFT)} & \multicolumn{2}{c}{Full Fine-tuning (FFT)} \\
\cmidrule(lr){2-3} \cmidrule(lr){4-5} \cmidrule(lr){6-7}
Organ & Dice & IoU & Dice & IoU & Dice & IoU \\
\midrule
Abdominal Wall     & 83.77 & 74.31 & $-$80.99 & $-$70.59 & \textbf{$+$84.54} & \textbf{$+$75.56} \\
Colon              & 68.93 & 57.83 & $-$64.95 & $-$52.46 & \textbf{$+$75.49} & \textbf{$+$64.50} \\
Inferior Mesenteric & \textbf{53.09} & \textbf{40.10} & $-$44.96 & $-$32.80 & $-$51.77 & $-$39.40 \\
Intestinal Veins   & 53.56 & 41.41 & $-$52.87 & $-$40.87 & \textbf{$+$55.10} & \textbf{$+$43.28} \\
Liver              & 55.01 & 45.78 & $-$51.16 & $-$39.62 & $-$50.66 & $-$41.84 \\
Pancreas           & 33.48 & 25.10 & $-$32.90 & $-$23.31 & \textbf{$+$41.17} & \textbf{$+$31.48} \\
Small Intestine    & 80.78 & 71.05 & $-$79.61 & $-$68.73 & \textbf{$+$83.95} & \textbf{$+$75.06} \\
Spleen             & \textbf{77.13} & \textbf{69.04} & $-$67.45 & $-$57.33 & $-$76.32 & $-$68.30 \\
Stomach            & 65.44 & 54.82 & $-$57.07 & $-$46.18 & \textbf{$+$67.23} & \textbf{$+$56.82} \\
Ureter             & 45.49 & 34.83 & $-$37.89 & $-$28.26 & \textbf{$+$50.12} & \textbf{$+$40.01} \\
Vesicular Glands   & 43.21 & 32.05 & $-$36.44 & $-$25.67 & \textbf{$+$50.36} & \textbf{$+$38.08} \\
\midrule
\textbf{Overall}   & 59.99 & 49.67 & $-$55.12 & $-$44.17 & \textbf{$+$62.43} & \textbf{$+$52.21} \\
\bottomrule
\end{tabular}
\end{table*}

\begin{figure*}[tbp]
    \centering
    \begin{subfigure}[b]{0.49\textwidth}
        \centering
        \includegraphics[width=\linewidth]{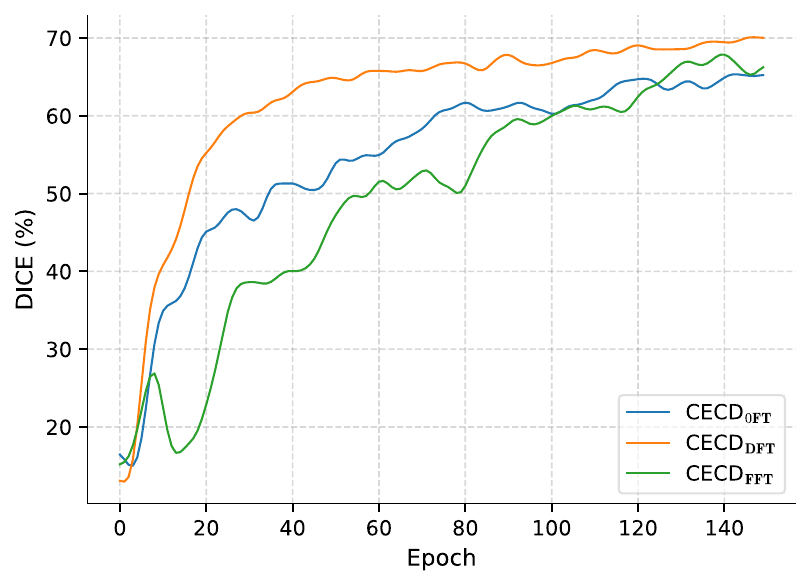}
        \caption{CECD}
        \label{fig:f1_cecd}
    \end{subfigure}
    \hfill
    \begin{subfigure}[b]{0.49\textwidth}
        \centering
        \includegraphics[width=\linewidth]{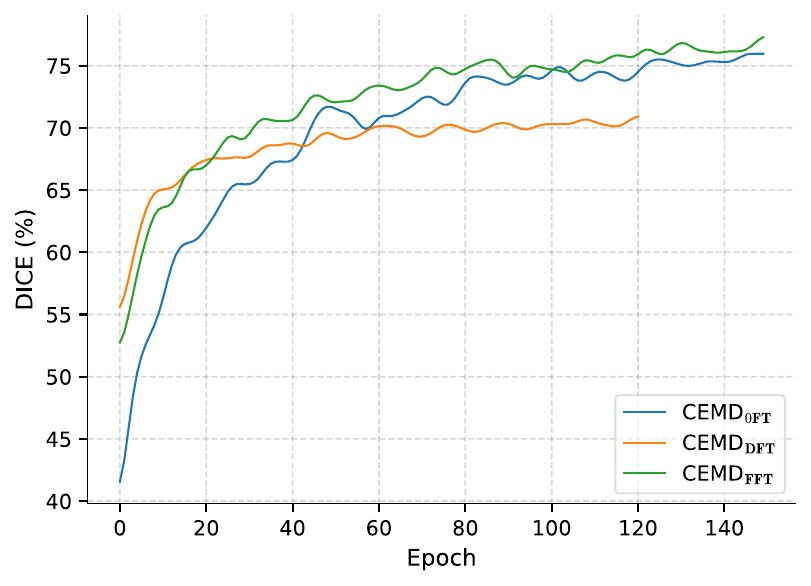}
        \caption{CEMD}
        \label{fig:f1_cemd}
    \end{subfigure}
    \caption{Comparison of DICE scores (\%) on the validation set across training epochs for training from scratch on the DSA dataset (0FT) and two transfer learning approaches: decoder fine-tuning (DFT), which only updates the decoder with the encoder frozen, and full fine-tuning (FFT), which updates the entire network. Both DFT and FFT use models pre-trained on CholecSeg8K.}
    \label{fig:f1}
\end{figure*}

Secondly, the results without any fine-tuning (0FT) present a significant insight, as the scores are in a similar range to the in-dataset evaluation on the DSA dataset in Figure~\ref{fig:cecd_cemd_dsa}. It proves the effectiveness of cross-dataset knowledge transfer in the surgical domain, notwithstanding the class incompatibilities.

Thirdly, the two fine-tuning approaches emphasize the importance of encoder adaptation to the target domain, demonstrating different outcomes where decoder-only fine-tuning (DFT) results in performance degradation and full fine-tuning (FFT) reflects performance improvements. Pre-trained models with a full fine-tuning approach (FFT) on the DSA dataset have reflected an increase mainly in the CEMD framework, with an increase of approximately 3\% in CEMD\textsubscript{FFT} in comparison to CEMD\textsubscript{0FT}. Figure~\ref{fig:dice_transfer_learning} delineates the influence of fine-tuning for CECD and CEMD frameworks by plotting the difference in dice scores. Refer to Tables~\ref{tab:cecd} and~\ref{tab:cemd} for detailed organ-wise dice and IoU scores. CEMD\textsubscript{FFT} achieves the highest overall dice score (62.4\%).

\subsubsection{Training Convergence}

Beyond peak performance gains, CEMD training accelerates convergence. Figure~\ref{fig:f1} presents the respective learning curves on validation sets at different training stages. CEMD\textsubscript{FFT} reached a 59.9\% dice score after only 50 epochs, comparable to the best checkpoint of CEMD\textsubscript{0FT} (60.0\%), which required training for the full duration. Similarly, CECD\textsubscript{FFT} demonstrated faster learning, reaching a 53.0\% dice score at 100 epochs versus 52.4\% for CECD\textsubscript{0FT}. This efficiency illustrates how transfer learning addresses the computational challenges in surgical image segmentation while improving overall performance.

The superior performance of full fine-tuning approaches compared to frozen encoder strategies indicates that surgical image segmentation benefits from end-to-end adaptation of feature hierarchies, allowing both encoder and decoder components to jointly optimize for task-specific anatomical representations. This finding challenges purely decoder-centric optimization approaches and suggests that holistic architectural adaptation yields superior results in cross-domain transfer scenarios.

Table~\ref{tab:summary} summarizes the overall dice scores on the DSA dataset for training from scratch and the best transfer learning configuration.

\begin{table}[btp]
\centering
\caption{Overall dice score (\%) comparison on the DSA dataset: training from scratch vs.\ best transfer learning configuration (FFT). $\Delta$ denotes the difference (FFT $-$ from scratch).}
\label{tab:summary}
\small
\begin{tabular}{lccc}
\toprule
Framework & From Scratch & FFT & $\Delta$ \\
\midrule
CECD & 59.3 & 59.4 & $+$0.1 \\
CEMD & 60.0 & 62.4 & $+$2.4 \\
\bottomrule
\end{tabular}
\end{table}

\section{Limitations}
Our study has several limitations that point to open problems in surgical segmentation. First, despite the consistent improvements offered by CEMD, class imbalance in surgical data remains a fundamental challenge that neither the architectural design nor transfer learning fully resolves. Under the best-performing configuration (CEMD\textsubscript{FFT}), organs with less than 3\% pixel share such as the Pancreas (dice 41.2\%), the Inferior Mesenteric Artery (51.8\%), and the Ureter (50.1\%) still lag substantially behind well-represented structures. Transfer learning primarily amplifies gains on classes that are already learnable, rather than closing the gap for the most underrepresented ones.

Second, segmentation performance depends on the visual and spatial characteristics of the masks and not only on pixel proportion. On CholecSeg8K, the Liver has the largest share in the training data but scores lower than the Fat class, despite comparable test-set proportions. This suggests that factors such as boundary clarity, texture consistency, and shape regularity play a significant role, and this dependence warrants further study.

Third, the transfer learning outcome is sensitive to the adaptation strategy. Decoder-only fine-tuning (DFT) consistently degrades performance relative to the no-fine-tuning baseline (CECD: $-$5.6\%, CEMD: $-$4.9\% overall dice), indicating that misalignment between a frozen encoder and an adapting decoder can be detrimental. Full-network fine-tuning is therefore necessary to realize transfer learning benefits, which increases the computational requirements for domain adaptation.

Fourth, our experiments use a single backbone architecture (Attention U-Net). The observed dynamics of encoder versus decoder adaptation may differ with other architectures, such as transformer-based encoders, limiting the generalizability of these specific findings.

Fifth, both datasets originate from single-center settings, and the generalizability of the reported transfer learning effects to multi-institutional or multi-device scenarios remains untested.

\section{Conclusion \& Future Work}
This work investigates architectural design choices and knowledge transfer strategies for multi-organ segmentation in laparoscopic surgery. Our findings validate the superiority of organ-specific decoder architectures (CEMD) over shared-decoder approaches (CECD) across two surgical datasets: on DSA, CEMD improves the overall dice score by approximately 3.5\% when trained from scratch, and on CholecSeg8K, by approximately 4\%, with the gain on CholecSeg8K concentrated on the underrepresented Gallbladder class ($+$27\%). These results corroborate and extend previous work \cite{tomar2025effective} to a second surgical domain.

Cross-surgical-domain transfer from CholecSeg8K to the DSA dataset proves effective: without any fine-tuning, the CEMD model achieves a dice score comparable to training from scratch on DSA (approximately 60\%), demonstrating that surgical anatomical knowledge transfers across procedures despite only partial class overlap. Full fine-tuning further improves performance, with CEMD\textsubscript{FFT} reaching the highest overall dice score of 62.4\% and converging substantially faster, matching the 150-epoch from-scratch performance in only 50 epochs.

The comparison of fine-tuning strategies reveals an important asymmetry: decoder-only fine-tuning degrades performance for both architectures, but CEMD is more resilient than CECD, suggesting that organ-specific decoders retain pre-trained knowledge better under domain shift. This implies that, in the CECD framework, the shared decoder is more tightly coupled to the encoder's feature distribution, making it more vulnerable when the encoder is frozen in a new domain.

Despite these improvements, class imbalance in surgical data remains the dominant bottleneck. Organs with very low pixel representation continue to underperform regardless of architecture or transfer strategy. Future work should explore targeted strategies for underrepresented classes, such as class-aware sampling, anatomy-specific augmentation, or loss re-weighting in conjunction with decoder-specific architectures. Additionally, extending the investigation to multiple surgical domains with joint multi-dataset training, evaluating alternative encoder architectures (e.g., vision transformers), and validating on multi-center data would strengthen the generalizability of these findings.

\section*{Acknowledgment}
This research has been funded by the Federal Ministry of Education and Research of Germany and the state of North-Rhine Westphalia as part of the Lamarr-Institute for Machine Learning and Artificial Intelligence, LAMARR22B. This research is also funded by the Deutsche Forschungsgemeinschaft (DFG, German Research Foundation) under Germany's Excellence Strategy---EXC 2070-390732324-PhenoRob.

\bibliographystyle{unsrtnat}
\bibliography{bibliography}

@article{zhou2025vision,
  title={Vision techniques for anatomical structures in laparoscopic surgery: a comprehensive review},
  author={Zhou, Ru and Wang, Dan and Zhang, Hanwei and Zhu, Ying and Zhang, Lijun and Chen, Tianxiang and Liao, Wenqiang and Ye, Zi},
  journal={Frontiers in Surgery},
  volume={12},
  pages={1557153},
  year={2025}
}

@article{rueckert2024methods,
  title={Methods and datasets for segmentation of minimally invasive surgical instruments in endoscopic images and videos: A review of the state of the art},
  author={Rueckert, Tobias and Rueckert, Daniel and Palm, Christoph},
  journal={Computers in Biology and Medicine},
  volume={169},
  pages={107929},
  year={2024},
  publisher={Elsevier}
}

@article{wu2025augmenting,
  title={Augmenting efficient real-time surgical instrument segmentation in video with point tracking and Segment Anything},
  author={Wu, Zijian and Schmidt, Adam and Kazanzides, Peter and Salcudean, Septimiu E},
  journal={Healthcare Technology Letters},
  volume={12},
  number={1},
  pages={e12111},
  year={2025},
  publisher={Wiley Online Library}
}

@article{wei2025segmatch,
  title={SegMatch: semi-supervised surgical instrument segmentation},
  author={Wei, Meng and Budd, Charlie and Garcia-Peraza-Herrera, Luis C and Dorent, Reuben and Shi, Miaojing and Vercauteren, Tom},
  journal={Scientific Reports},
  volume={15},
  number={1},
  pages={14042},
  year={2025},
  publisher={Nature Publishing Group UK London}
}

@article{ahmed2024deep,
  title={Deep learning for surgical instrument recognition and segmentation in robotic-assisted surgeries: a systematic review},
  author={Ahmed, Fatimaelzahraa Ali and Yousef, Mahmoud and Ahmed, Mariam Ali and Ali, Hasan Omar and Mahboob, Anns and Ali, Hazrat and Shah, Zubair and Aboumarzouk, Omar and Al Ansari, Abdulla and Balakrishnan, Shidin},
  journal={Artificial Intelligence Review},
  volume={58},
  number={1},
  pages={1},
  year={2024},
  publisher={Springer}
}

@article{tf1,
title = {Transfer learning in medical image segmentation: New insights from analysis of the dynamics of model parameters and learned representations},
journal = {Artificial Intelligence in Medicine},
volume = {116},
pages = {102078},
year = {2021},
issn = {0933-3657},
author = {Davood Karimi and Simon K. Warfield and Ali Gholipour},
}

@article{twinanda2016endonet,
  title={Endonet: a deep architecture for recognition tasks on laparoscopic videos},
  author={Twinanda, Andru P and Shehata, Sherif and Mutter, Didier and Marescaux, Jacques and De Mathelin, Michel and Padoy, Nicolas},
  journal={IEEE transactions on medical imaging},
  volume={36},
  number={1},
  pages={86--97},
  year={2016},
  publisher={IEEE}
}

@inbook{tf2,
author = {Raghu, Maithra and Zhang, Chiyuan and Kleinberg, Jon and Bengio, Samy},
title = {Transfusion: understanding transfer learning for medical imaging},
year = {2019},
publisher = {Curran Associates Inc.},
address = {Red Hook, NY, USA},
booktitle = {Proceedings of the 33rd International Conference on Neural Information Processing Systems},
articleno = {301},
numpages = {11}
}

@unknown{tf3,
author = {Karimi, Davood and Warfield, Simon and Gholipour, Ali},
year = {2020},
month = {05},
pages = {},
title = {Critical Assessment of Transfer Learning for Medical Image Segmentation with Fully Convolutional Neural Networks},
doi = {10.48550/arXiv.2006.00356}
}

@article {intro1,
	author = {Kolbinger, Fiona R. and Rinner, Franziska M. and Jenke, Alexander C. and Carstens, Matthias and Leger, Stefan and others},
	title = {Anatomy Segmentation in Laparoscopic Surgery: Comparison of Machine Learning and Human Expertise},
	elocation-id = {2022.11.11.22282215},
	year = {2023},
	doi = {10.1101/2022.11.11.22282215},
	publisher = {Cold Spring Harbor Laboratory Press},
	journal = {medRxiv}
}

@article{intro2,
  title={A review of deep learning based methods for medical image multi-organ segmentation},
  author={Fu, Yabo and Lei, Yang and Wang, Tonghe and Curran, Walter J and Liu, Tian and Yang, Xiaofeng},
  journal={Physica Medica},
  volume={85},
  pages={107--122},
  year={2021},
  publisher={Elsevier}
}

@article{hong2020cholecseg8k,
  title={Cholecseg8k: a semantic segmentation dataset for laparoscopic cholecystectomy based on cholec80},
  author={Hong, W-Y and Kao, C-L and Kuo, Y-H and Wang, J-R and Chang, W-L and Shih, C-S},
  journal={arXiv preprint arXiv:2012.12453},
  year={2020}
}

@article{ch8k_split,
title = {Deep learning for semantic segmentation of organs and tissues in laparoscopic surgery},
author = {Paul Maria Scheikl and Stefan Laschewski and Anna Kisilenko and Tornike Davitashvili and Benjamin Müller and Manuela Capek and Beat P. Müller-Stich and Martin Wagner and Franziska Mathis-Ullrich},
pages = {20200016},
volume = {6},
number = {1},
journal = {Current Directions in Biomedical Engineering},
year = {2020},
lastchecked = {2025-04-10}
}

@article{article_dsad,
author = {Carstens, Matthias and Rinner, Franziska and Bodenstedt, Sebastian and Jenke, Alexander and Weitz, Jürgen and others},
year = {2023},
month = {01},
pages = {},
title = {The Dresden Surgical Anatomy Dataset for Abdominal Organ Segmentation in Surgical Data Science},
volume = {10},
journal = {Scientific Data},
doi = {10.1038/s41597-022-01719-2}
}

@inproceedings{
tomar2025effective,
title={Effective Disjoint Representational Learning for Anatomical Segmentation},
author={Priya Tomar and Aditya Parikh and Philipp Feodorovici and Jan Arensmeyer and Hanno Matthaei and Christian Bauckhage and Helen Schneider and Rafet Sifa},
booktitle={Medical Imaging with Deep Learning},
year={2025},
}

@article{aunet,
  author       = {Ozan Oktay and
                  Jo Schlemper and
                  Lo{\"{\i}}c Le Folgoc and
                  Matthew C. H. Lee and
                  Mattias P. Heinrich and
                  others},
  title        = {Attention U-Net: Learning Where to Look for the Pancreas},
  journal      = {CoRR},
  volume       = {abs/1804.03999},
  year         = {2018},
  eprinttype    = {arXiv},
  eprint       = {1804.03999},
}

@INPROCEEDINGS{para11,
  author={D. K. Venkatesh and D. Rivoir and M. Pfeiffer and F. Kolbinger and S. Speidel},
  booktitle={2025 IEEE/CVF Winter Conference on Applications of Computer Vision (WACV)},
  title={Data Augmentation for Surgical Scene Segmentation with Anatomy-Aware Diffusion Models},
  year={2025},
  pages={2280--2290},
  doi={10.1109/WACV61041.2025.00228}
}

@article{para13,
title = {MOSMOS: Multi-organ segmentation facilitated by medical report supervision},
journal = {Biomedical Signal Processing and Control},
volume = {106},
pages = {107743},
year = {2025},
issn = {1746-8094},
author = {Weiwei Tian and Xinyu Huang and Junlin Hou and Caiyue Ren and Longquan Jiang and Rui-Wei Zhao and Gang Jin and Yuejie Zhang and Daoying Geng}
}

@misc{para14,
      title={Towards more precise automatic analysis: a comprehensive survey of deep learning-based multi-organ segmentation}, 
      author={Xiaoyu Liu and Linhao Qu and Ziyue Xie and Jiayue Zhao and Yonghong Shi and Zhijian Song},
      year={2023},
      eprint={2303.00232},
      archivePrefix={arXiv},
      primaryClass={eess.IV},
}

@article{para22,
author = {Kaur, Harinder and Anttal, Navjot and Neeru, Nirvair},
year = {2022},
month = {04},
pages = {102223},
title = {Evolution of Multiorgan Segmentation Techniques from Traditional to Deep Learning in Abdominal CT Images – A Systematic Review},
volume = {73},
journal = {Displays},
doi = {10.1016/j.displa.2022.102223}
}

@article{dsad1,
  title={One model to use them all: training a segmentation model with complementary datasets},
  author={Jenke, Alexander C and Bodenstedt, Sebastian and Kolbinger, Fiona R and Distler, Marius and Weitz, J{\"u}rgen and Speidel, Stefanie},
  journal={International journal of computer assisted radiology and surgery},
  volume={19},
  number={6},
  pages={1233--1241},
  year={2024},
  publisher={Springer}
}

@inproceedings{kolbinger2024strategies,
  title={Strategies to improve real-world applicability of laparoscopic anatomy segmentation models},
  author={Kolbinger, Fiona R and He, Jiangpeng and Ma, Jinge and Zhu, Fengqing},
  booktitle={Proceedings of the IEEE/CVF Conference on Computer Vision and Pattern Recognition},
  pages={2275--2284},
  year={2024}
}

@inproceedings{dsad2,
title={Efficient Anatomy Segmentation in Laparoscopic Surgery using Multi-Teacher Knowledge Distillation},
author={Lennart Maack and Finn Behrendt and Debayan Bhattacharya and Sarah Latus and Alexander Schlaefer},
booktitle={Medical Imaging with Deep Learning},
year={2024},
}

@article{qi2025advancements,
  title={Advancements and Challenges in Medical Image Segmentation: A Comprehensive Survey},
  author={Qi, Guanqiu and Zhu, Zhiqin and Li, Ke and Xiao, Han},
  journal={Sensors and AI},
  pages={3--29},
  year={2025}
}

@article{urrea2024improving,
  title={Improving Surgical scene semantic segmentation through a deep learning architecture with attention to class imbalance},
  author={Urrea, Claudio and Garcia-Garcia, Yainet and Kern, John},
  journal={Biomedicines},
  volume={12},
  number={6},
  pages={1309},
  year={2024},
  publisher={MDPI}
}

@article{liu2024towards,
  title={Towards more precise automatic analysis: a systematic review of deep learning-based multi-organ segmentation},
  author={Liu, Xiaoyu and Qu, Linhao and Xie, Ziyue and Zhao, Jiayue and Shi, Yonghong and Song, Zhijian},
  journal={BioMedical Engineering OnLine},
  volume={23},
  number={1},
  pages={52},
  year={2024},
  publisher={Springer}
}

@article{chen2021transunet,
  title={Transunet: Transformers make strong encoders for medical image segmentation},
  author={Chen, Jieneng and Lu, Yongyi and Yu, Qihang and Luo, Xiangde and Adeli, Ehsan and Wang, Yan and Lu, Le and Yuille, Alan L and Zhou, Yuyin},
  journal={arXiv preprint arXiv:2102.04306},
  year={2021}
}

@article{wang2025efficient,
  title={Efficient Generative-Adversarial U-Net for Multi-Organ Medical Image Segmentation},
  author={Wang, Haoran and Wu, Gengshen and Liu, Yi},
  journal={Journal of Imaging},
  volume={11},
  number={1},
  pages={19},
  year={2025}
}

@inproceedings{alapatt2024jumpstarting,
  title={Jumpstarting Surgical Computer Vision},
  author={Alapatt, Deepak and Murali, Aditya and Srivastav, Vinkle and {AI4SafeChole Consortium} and Mascagni, Pietro and Padoy, Nicolas},
  booktitle={Medical Image Computing and Computer Assisted Intervention -- MICCAI 2024},
  pages={328--338},
  year={2024},
  publisher={Springer}
}

@article{jaspers2025surgenet,
  title={Scaling up self-supervised learning for improved surgical foundation models},
  author={Jaspers, Tim J.M. and de Jong, Ronald L.P.D. and Li, Yiping and Kusters, Carolus H.J. and Bakker, Franciscus H.A. and van Jaarsveld, Romy C. and Kuiper, Gino M. and van Hillegersberg, Richard and Ruurda, Jelle P. and Brinkman, Willem M. and Pluim, Josien P.W. and de With, Peter H.N. and Breeuwer, Marcel and Al Khalil, Yasmina and van der Sommen, Fons},
  journal={Medical Image Analysis},
  volume={108},
  pages={103873},
  year={2026},
  publisher={Elsevier}
}

@inproceedings{xu2025balance,
  title={A Unified Loss for Handling Inter-Class and Intra-Class Imbalance in Medical Image Segmentation},
  author={Xu, Fei and Yang, Fan and Li, Xinghui and Zhang, Xiaofeng},
  booktitle={Proceedings of the AAAI Conference on Artificial Intelligence},
  volume={39},
  number={8},
  pages={8842--8850},
  year={2025}
}

\end{document}